\documentclass[11pt,a4paper]{article}
\usepackage[hyperref]{emnlp-ijcnlp-2019}
\usepackage{times}
\usepackage{latexsym}
\usepackage{xcolor}
\usepackage{amsmath, amssymb, amsthm, enumerate, framed, graphicx}
\usepackage[lofdepth,lotdepth]{subfig}
\usepackage[normalem]{ulem}
\usepackage{booktabs}
\usepackage{multirow, makecell}
\usepackage{url}
\usepackage{arydshln}
\usepackage{xcolor}
\usepackage{bm}

\aclfinalcopy 

\newcommand{\xv}{\mathbf{x}}
\newcommand{\yv}{\mathbf{y}}
\newcommand{\zv}{\mathbf{z}}

\newcommand{\argmax}{\operatornamewithlimits{argmax}}

\definecolor{mypink}{cmyk}{0, 0.7808, 0.4429, 0.1412}
\definecolor{mypurple}{rgb}{0.57, 0.36, 0.51}
\definecolor{myblue}{rgb}{0.2, 0.2, 0.6}

\title{FlowSeq:  Non-Autoregressive Conditional \\ Sequence Generation with Generative Flow}

\author{Xuezhe Ma$^{*,1}$ $\quad$ Chunting Zhou\thanks{$\,\,$ Equal contribution, in alphabetical order.}$\,\,^{,1}$ $\quad$ Xian Li$^2$ $\quad$ Graham Neubig$^1$ $\quad$ Eduard Hovy$^1$ \\
  $^1$Language Technologies Institute, Carnegie Mellon University $\,$\\ $^2$Facebook AI \\
     {\tt \{xuezhem, chuntinz, gneubig, ehovy\}@cs.cmu.edu ~xianl@fb.com} \\
}

\date{}

\begin{document}
\maketitle
\begin{abstract}
Most sequence-to-sequence (seq2seq) models are \emph{autoregressive}; they generate each token by conditioning on previously generated tokens.
In contrast, non-autoregressive seq2seq models generate all tokens in one pass, which leads to increased efficiency through parallel processing on hardware such as GPUs. 
However, directly modeling the joint distribution of all tokens simultaneously is challenging, and even with increasingly complex model structures accuracy lags significantly behind autoregressive models.
In this paper, we propose a simple, efficient, and effective model for non-autoregressive sequence generation using latent variable models.
Specifically, we turn to generative flow, an elegant technique to model complex distributions using neural networks, and design several layers of flow tailored for modeling the conditional density of sequential latent variables.
We evaluate this model on three neural machine translation (NMT) benchmark datasets, achieving comparable performance with state-of-the-art non-autoregressive NMT models and almost constant decoding time w.r.t the sequence length.\footnote{\url{https://github.com/XuezheMax/flowseq}}
\end{abstract}

\section{Introduction}

Neural sequence-to-sequence (seq2seq) models \citep{bahdanau2014neural,rush2015neural,vinyals2015show,vaswani2017attention} generate an output sequence $\mathbf{y} = \{y_1, \ldots, y_T\}$ given an input sequence $\mathbf{x} = \{x_1, \ldots, x_{T'}\}$ using conditional probabilities $P_\theta(\mathbf{y}|\mathbf{x})$ predicted by neural networks (parameterized by $\theta$).

\begin{figure}[tb]
  \centering
  \includegraphics[scale=0.61]{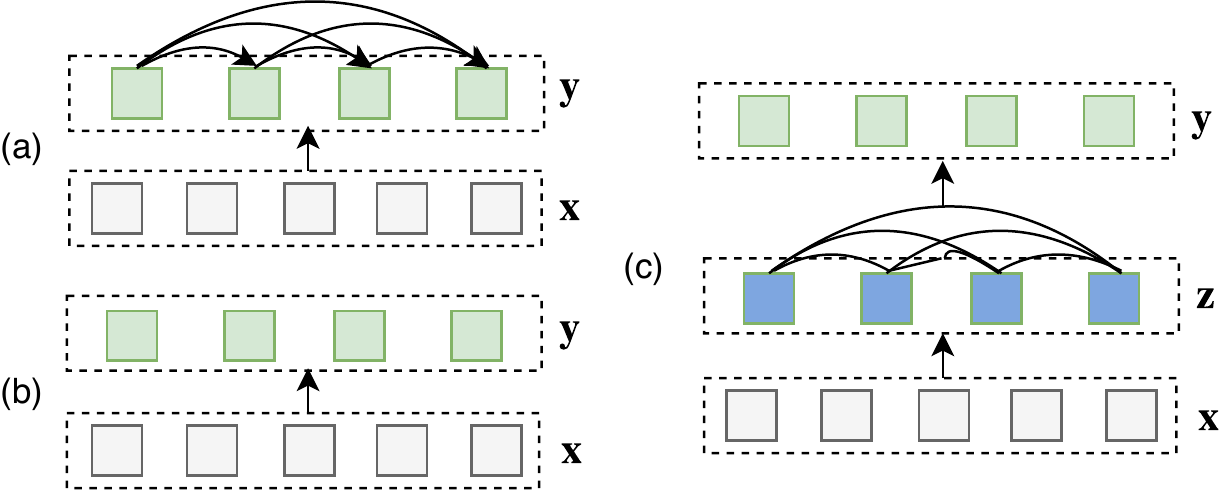}
  \caption{(a) Autoregressive (b) non-autoregressive and (c) our proposed sequence generation models. $\mathbf{x}$ is the source, $\mathbf{y}$ is the target, and $\mathbf{z}$ are latent variables.} \label{fig:diagram}
  \vspace{-4mm}
\end{figure}

Most seq2seq models are \emph{autoregressive}, meaning that they factorize the joint probability of the output sequence given the input sequence $P_\theta(\mathbf{y}|\mathbf{x})$ into the product of probabilities over the next token in the sequence given the input sequence and previously generated tokens:

\begin{equation}\label{eq:autoregress}
P_\theta(\mathbf{y}|\mathbf{x}) = \prod\limits_{t=1}^{T} P_\theta(y_{t} | y_{<t}, \mathbf{x}).
\end{equation}

Each factor, $P_\theta(y_{t} | y_{<t}, \mathbf{x})$, can be implemented by function approximators such as RNNs~\citep{bahdanau2014neural} and Transformers~\citep{vaswani2017attention}.
This factorization takes the complicated problem of \emph{joint estimation} over an exponentially large output space of outputs $\mathbf{y}$, and turns it into a \emph{sequence of tractable multi-class classification problems} predicting $y_t$ given the previous words, allowing for simple maximum log-likelihood training.
However, this assumption of left-to-right factorization may be sub-optimal from a modeling perspective \cite{gu2019insertion,stern2019insertion}, and generation of outputs must be done through a linear left-to-right pass through the output tokens using beam search, which is not easily parallelizable on hardware such as GPUs.

Recently, there has been work on non-autoregressive sequence generation for neural machine translation (NMT; \citet{gu2018non,lee2018deterministic,constant2019}) and language modeling~\citep{ziegler2019latent}.
Non-autoregressive models attempt to model the joint distribution $P_\theta(\mathbf{y}|\mathbf{x})$ directly, decoupling the dependencies of decoding history during generation.
A na\"ive solution is to assume that each token of the target sequence is independent given the input:
\begin{equation}\label{eq:baseline}
P_\theta(\yv|\xv) = \prod\limits_{t=1}^{T} P_\theta(y_{t} | \mathbf{x}).
\vspace{-1mm}
\end{equation}
Unfortunately, the performance of this simple model falls far behind autoregressive models, as seq2seq tasks usually do have strong conditional dependencies between output variables~\citep{gu2018non}. This problem can be mitigated by introducing a latent variable $\zv$ to model these conditional dependencies:
\begin{equation}\label{eq:non_auto}
P_{\theta}(\yv| \mathbf{x}) = \int_{\zv} P_{\theta}(\mathbf{y}|\zv, \mathbf{x}) p_{\theta}(\zv|\mathbf{x}) d\zv,
\vspace{-2mm}
\end{equation}
where $p_{\theta}(\zv|\mathbf{x})$ is the prior distribution over latent $\zv$ and $P_{\theta}(\yv|\zv, \xv)$ is the ``generative'' distribution (a.k.a decoder).
Non-autoregressive generation can be achieved by the following independence assumption in the decoding process:
\begin{equation}\label{eq:decoding}
P_{\theta}(\yv|\zv, \mathbf{x}) = \prod\limits_{t=1}^{T} P_\theta(y_{t} | \zv, \mathbf{x}).
\vspace{-2mm}
\end{equation}
\citet{gu2018non} proposed a $\zv$ representing fertility scores specifying the number of output words each input word generates, significantly improving the performance over Eq.~\eqref{eq:baseline}. 
But the performance still falls behind state-of-the-art autoregressive models due to the limited expressiveness of fertility to model the interdependence between words in $\textbf{y}$.

In this paper, we propose a simple, effective, and efficient model, \textbf{FlowSeq}, which models expressive prior distribution $p_{\theta}(\zv|\mathbf{x})$ using a powerful mathematical framework called generative flow \cite{rezende2015variational}.
This framework can elegantly model complex distributions, and has obtained remarkable success in modeling continuous data such as images and speech through efficient density estimation and sampling \cite{kingma2018glow,prenger2019waveglow,ma2019macow}.
Based on this, we posit that generative flow also has potential to introduce more meaningful latent variables $\zv$ in the non-autoregressive generation in Eq.~\eqref{eq:non_auto}.

FlowSeq is a \emph{flow-based sequence-to-sequence} model, which is (to our knowledge) the first non-autoregressive seq2seq model utilizing generative flows.
It allows for efficient parallel decoding while modeling the joint distribution of the output sequence.
Experimentally, on three benchmark datasets for machine translation -- WMT2014, WMT2016 and IWSLT-2014, 
FlowSeq achieves comparable performance with state-of-the-art non-autoregressive models, and almost constant decoding time w.r.t.~the sequence length compared to a typical left-to-right Transformer model, which is super-linear.

\section{Background}
As noted above, incorporating expressive latent variables $\zv$ is essential to decouple the dependencies between tokens in the target sequence in non-autoregressive models.
However, in order to model all of the complexities of sequence generation to the point that we can read off all of the words in the output in an independent fashion (as in Eq.~\eqref{eq:decoding}), the prior distribution $p_{\theta}(\zv|\mathbf{x})$ will necessarily be quite complex.
In this section, we describe generative flows \cite{rezende2015variational}, an effective method for arbitrary modeling of complicated distributions, before describing how we apply them to sequence-to-sequence generation in \S\ref{sec:flowseq}.

\subsection{Flow-based Generative Models}
\label{subsec:flow}
Put simply, flow-based generative models work by transforming a simple distribution (e.g.~a simple Gaussian) into a complex one (e.g.~the complex prior distribution over $\zv$ that we want to model) through a chain of invertible transformations.

Formally, a set of latent variables $\bm{\upsilon} \in \Upsilon$ are introduced with a simple prior distribution $p_{\Upsilon}(\upsilon)$.
We then define a bijection function $f: \mathcal{Z} \rightarrow \Upsilon$ (with $g = f^{-1}$), whereby we can define a generative process over variables $\zv$:
\begin{equation}
\label{eq:generation}
\begin{array}{rcl}
\bm{\upsilon} & \sim & p_{\Upsilon}(\bm{\upsilon}) \\
\zv & = & g_{\theta}(\bm{\upsilon}).
\end{array}
\vspace{-2mm}
\end{equation}
An important insight behind flow-based models is that given this bijection function, the change of variable formula defines the model distribution on $\zv \in \mathcal{Z}$ by:
\begin{equation}\label{eq:cvf}
p_{\theta}(\zv) = p_{\Upsilon}(f_{\theta}(\zv))\left| \det(\frac{\partial f_{\theta}(\zv)}{\partial \zv})\right|.
\vspace{-2mm}
\end{equation}
Here $\frac{\partial f_{\theta}(\zv)}{\partial \zv}$ is the Jacobian matrix of $f_{\theta}$ at $\zv$.

Eq.~\eqref{eq:cvf} provides a way to calculate the (complex) density of $\zv$ by calculating the (simple) density of $\upsilon$ and the Jacobian of the transformation from $\zv$ to $\upsilon$.
For efficiency purposes, flow-based models generally use certain types of transformations $f_{\theta}$ where both the inverse functions $g_{\theta}$ and the Jacobian determinants are tractable to compute.
A stacked sequence of such invertible transformations is also called a (normalizing) \emph{flow}~\citep{rezende2015variational}:
\begin{displaymath}
\zv \underset{g_1}{\overset{f_1}{\longleftrightarrow}} H_1 \underset{g_2}{\overset{f_2}{\longleftrightarrow}} H2 \underset{g_3}{\overset{f_3}{\longleftrightarrow}} \cdots \underset{g_K}{\overset{f_K}{\longleftrightarrow}} \bm{\upsilon},
\end{displaymath}
where $f = f_1 \circ f_2 \circ \cdots \circ f_K$ is a flow of $K$ transformations (omitting $\theta$s for brevity).

\subsection{Variational Inference and Training}
\label{subsec:vlf}
In the context of maximal likelihood estimation (MLE), we wish to minimize the negative log-likelihood of the parameters:
\begin{equation}\label{eq:mle}
\min\limits_{\theta \in \Theta} \frac{1}{N} \sum\limits_{i=1}^{N} -\log P_{\theta}(\yv^i | \xv^i),
\end{equation}
where $D=\{(\xv^i, \yv^i)\}_{i=1}^{N}$ is the set of training data.
However, the likelihood $P_{\theta}(\yv | \xv)$ after marginalizing out latent variables $\zv$ (LHS in Eq.~\eqref{eq:non_auto}) is intractable to compute or differentiate directly.
Variational inference~\citep{wainwright2008graphical} provides a solution by introducing a parametric \emph{inference model} $q_{\phi}(\zv|\yv, \xv)$ (a.k.a posterior) which is then used to approximate this integral by sampling individual examples of $\zv$.
These models then optimize the \emph{evidence lower bound} (ELBO), which considers both the ``reconstruction error'' $\log P_\theta(\yv|\zv,\xv)$ and KL-divergence between the posterior and the prior:
\begin{align}\label{eq:elbo}
\log P_{\theta}(\yv | \xv) & \geq \mathrm{E}_{q_{\phi} (\zv|\yv, \xv)} [\log P_{\theta}(\yv|\zv, \xv)] \nonumber \\
 & - \mathrm{KL}(q_{\phi} (\zv|\yv, \xv) || p_{\theta}(\zv|\xv)).
\end{align}
Both inference model $\phi$ and decoder $\theta$ parameters are optimized according to this objective.

\begin{figure*}[tb]
  \centering
  \includegraphics[scale=0.8]{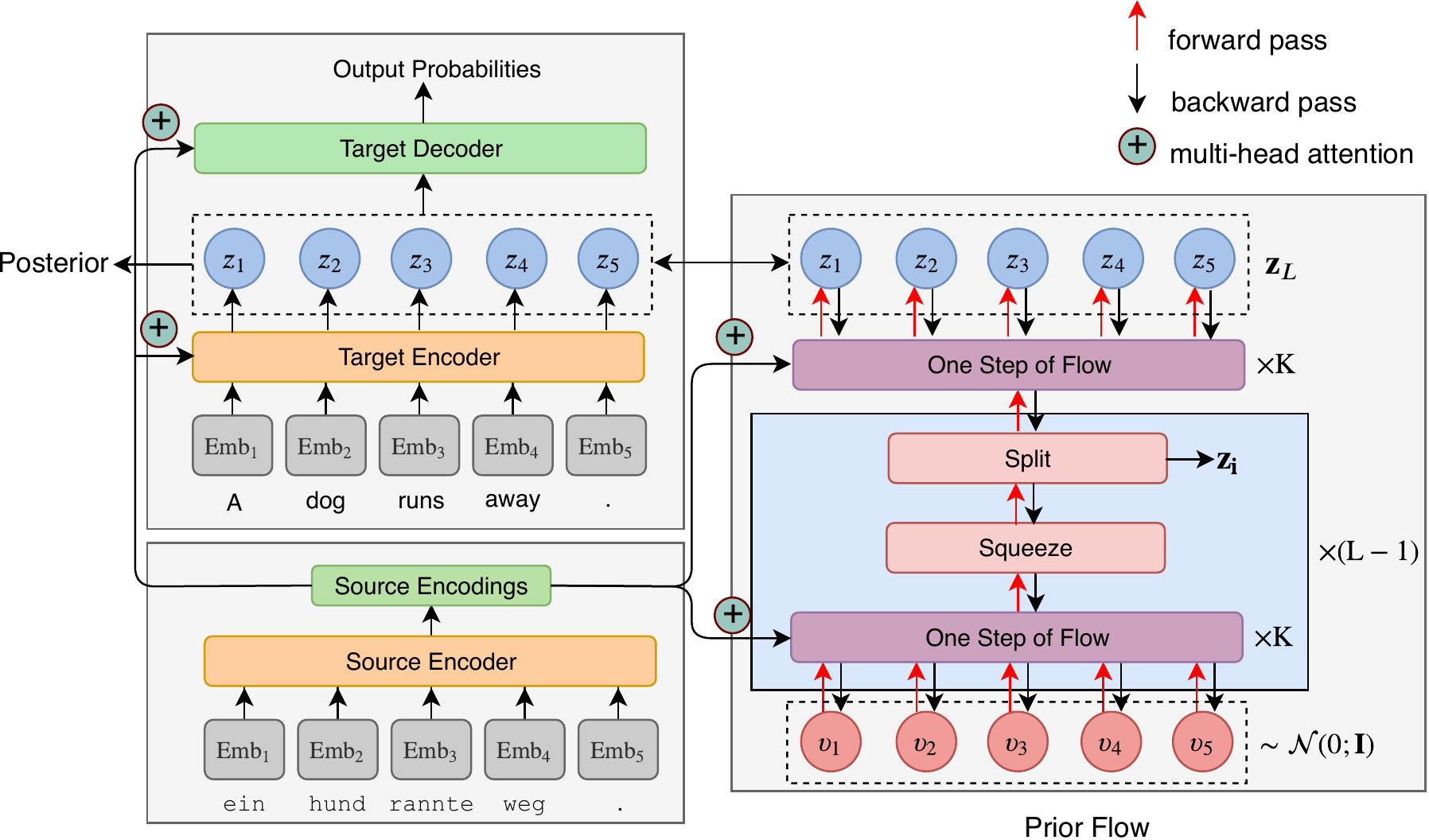}
  \caption{Neural architecture of FlowSeq, including the encoder, the decoder and the posterior networks, together with the multi-scale architecture of the prior flow. The architecture of each flow step is in Figure~\ref{fig:flowstep}. 
  } \label{fig:model}
  \vspace{-4mm}
\end{figure*}

\section{FlowSeq}
\label{sec:flowseq}

We first overview FlowSeq's architecture (shown in Figure~\ref{fig:model}) and training process here before detailing each component in following sections.
Similarly to classic seq2seq models, at both training and test time FlowSeq first reads the whole input sequence $\xv$ and calculates a vector for each word in the sequence, the source encoding.

At training time, FlowSeq's parameters are learned using a variational training paradigm overviewed in \S\ref{subsec:vlf}.
First, we draw samples of latent codes $\zv$ from the current posterior $q_{\phi} (\zv|\yv, \xv)$. 
Next, we feed $\zv$ together with source encodings into the decoder network and the prior flow to compute the probabilities of $P_{\theta}(\yv|\zv, \xv)$ and $p_{\theta}(\zv|\xv)$ for optimizing the ELBO (Eq.~\eqref{eq:elbo}).

At test time, generation is performed by first sampling a latent code $\zv$ from the prior flow by executing the generative process defined in Eq.~\eqref{eq:generation}.
In this step, the source encodings produced from the encoder are used as conditional inputs.
Then the decoder receives both the sampled latent code $\zv$ and the source encoder outputs to generate the target sequence $\yv$ from $P_{\theta}(\yv|\zv, \xv)$.

\subsection{Source Encoder}
The source encoder encodes the source sequences into hidden representations, which are used in computing attention when generating latent variables in the posterior network and prior network as well as the cross-attention with decoder. 
Any standard neural sequence model can be used as its encoder, including RNNs \citep{bahdanau2014neural} or Transformers \cite{vaswani2017attention}.

\subsection{Posterior}
\paragraph{Generation of Latent Variables.} The latent variables $\zv$ are represented as a sequence of continuous random vectors $\zv=\{\zv_1, \ldots, \zv_T\}$ with the same length as the target sequence $\mathbf{y}$. Each $\zv_t$ is a $d_{\mathrm{z}}$-dimensional vector, where $d_{\mathrm{z}}$ is the dimension of the latent space. The posterior distribution $q_{\phi} (\zv|\yv, \xv)$ models each $\zv_t$ as a diagonal Gaussian with learned mean and variance:
\begin{equation}\label{eq:posterior}
q_{\phi} (\zv|\yv, \xv) = \prod\limits_{t=1}^{T} \mathcal{N}(\zv_t| \mu_{t}(\xv, \yv), \sigma_{t}^{2}(\xv, \yv))
\end{equation}
where $\mu_{t}(\cdot)$ and $\sigma_{t}(\cdot)$ are neural networks such as RNNs or Transformers.

\paragraph{Zero initialization.} While we perform standard random initialization for most layers of the network, we initialize the last linear transforms that generate the $\mu$ and $\log \sigma^2$ values with zeros.
This ensures that the posterior distribution as a simple normal distribution, which we found helps train very deep generative flows more stably.

\paragraph{Token Dropout.} 
The motivation of introducing the latent variable $\zv$ into the model is to model the uncertainty in the generative process.
Thus, it is preferable that $\zv$ capture contextual interdependence between tokens in $\yv$.
However, there is an obvious local optimum where the posterior network generates a latent vector $\zv_t$ that only encodes the information about the corresponding target token $y_t$, and the decoder simply generates the ``correct'' token at each step $t$ with $\zv_t$ as input.
In this case, FlowSeq reduces to the baseline model in Eq.~\eqref{eq:baseline}.
To escape this undesired local optimum, we apply token-level dropout to randomly drop an entire token when calculating the posterior, to ensure the model also has to learn how to use contextual information.
This technique is similar to the ``masked language model'' in previous studies~\citep{melamud-etal-2016-context2vec,devlin2018bert,P18-1130}.

\vspace{-2mm}
\subsection{Decoder}
As the decoder, we take the latent sequence $\zv$ as input, run it through several layers of a neural sequence model such as a Transformer, then directly predict the output tokens in $\yv$ individually and independently.
Notably, unlike standard seq2seq decoders,
we do not perform causal masking to prevent attending to future tokens,
making the model fully non-autoregressive.

\begin{figure*}[tb]
  \centering
  \includegraphics[scale=0.8]{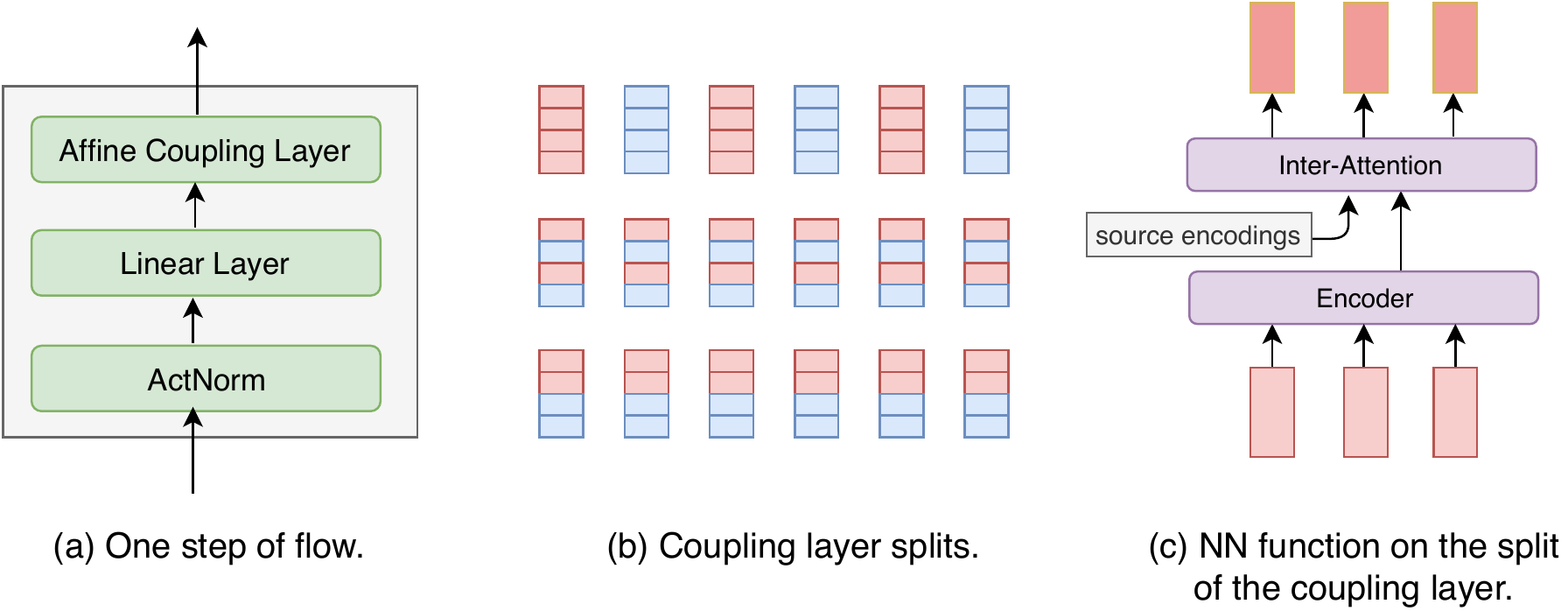}
  \vspace{-2mm}
  \caption{(a) The architecture of one step of our flow.
  (b) The visualization of three split pattern for coupling layers, where the red color denotes $\zv_a$ and the blue color denotes $zv_b$.
  (c) The attention-based architecture of the NN function in coupling layers.} 
  \label{fig:flowstep}
  \vspace{-5mm}
\end{figure*}

\subsection{Flow Architecture for Prior}
The flow architecture is based on Glow~\citep{kingma2018glow}.
It consists of a series of steps of flow, combined in a multi-scale architecture (see Figure~\ref{fig:model}.)
Each step of flow consists three types of elementary flows -- actnorm, invertible multi-head linear, and coupling.
Note that all three functions are invertible and conducive to calculation of log determinants (details in Appendix~\ref{appendix:flow}).

\paragraph{Actnorm.} The activation normalization layer (actnorm; \citet{kingma2018glow}) is an alternative for batch normalization~\citep{ioffe2015batch}, that has mainly been used in the context of image data to alleviate problems in model training. 
Actnorm performs an affine transformation of the activations using a scale and bias parameter per feature for sequences:

{\small
\begin{equation}\label{eq:actnorm}
\zv_{t}' = \mathbf{s} \odot \zv_{t} + \mathbf{b}.
\end{equation}
}
Both $\zv$ and $\zv'$ are tensors of shape $[T\times d_{\mathrm{z}}]$ with time dimension $t$ and feature dimension $d_{\mathrm{z}}$.
The parameters are initialized such that over each feature $\zv_{t}'$ has zero mean and unit variance given an initial mini-batch of data. 

\paragraph{Invertible Multi-head Linear Layers.}
To incorporate general permutations of variables along the feature dimension to ensure that each dimension can affect every other ones after a sufficient number of steps of flow, \citet{kingma2018glow} proposed a trainable invertible $1\times1$ convolution layer for 2D images.
It is straightforward to apply similar transformations to sequential data:

\vspace{-1mm}
{\small
\begin{equation}\label{eq:linear}
\zv_{t}' = \zv_{t} \mathbf{W},
\end{equation}
}
where $\mathbf{W}$ is the weight matrix of shape $[d_{\mathrm{z}} \times d_{\mathrm{z}}]$.
The log-determinant of this transformation is:

\vspace{-2mm}
{\small
\begin{displaymath}
\log \left| \mathrm{det} \left( \frac{\partial \mathrm{linear}(\zv; \mathbf{W})}{\partial \zv}\right) \right| = T \cdot \log |\mathrm{det} (\mathbf{W})|
\end{displaymath}
}
The cost of computing $\mathrm{det}(\mathbf{W})$ is $O(d_{\mathrm{z}}^3)$.

Unfortunately, $d_{\mathrm{z}}$ in Seq2Seq generation is commonly large, e.g. $512$, significantly slowing down the model for computing $\mathrm{det}(\mathbf{W})$. 
To apply this to sequence generation, we propose a multi-head invertible linear layer, which first splits each $d_{\mathrm{z}}$-dimensional feature vector into $h$ heads with dimension $d_h = d_{\mathrm{z}}/h$.
Then the linear transformation in \eqref{eq:linear} is applied to each head, with $d_h\times d_h$ weight matrix $\mathbf{W}$, significantly reducing the dimension.
For splitting of heads, one step of flow contains one linear layer with either row-major or column-major splitting format, and these steps with different linear layers are composed in an alternating pattern.

\paragraph{Affine Coupling Layers.} 
To model interdependence across time steps, we use affine coupling layers~\citep{dinh2016density}:
\begin{displaymath}
\begin{array}{rcl}
\zv_a, \zv_b & = & \mathrm{split}(\zv) \\
\zv_a' & = & \zv_a \\
\zv_b' & = & \mathrm{s}(\zv_a, \xv) \odot \zv_b + \mathrm{b}(\zv_a, \xv) \\
\zv' & = & \mathrm{concat}(\zv_a', \zv_b'),
\end{array}
\end{displaymath}
where $\mathrm{s}(\zv_a, \xv)$ and $\mathrm{b}(\zv_a, \xv)$ are outputs of two neural networks with $\zv_a$ and $\xv$ as input.
These are shown in Figure \ref{fig:flowstep} (c).
In experiments, we implement $\mathrm{s}(\cdot)$ and $\mathrm{b}(\cdot)$ with one Transformer decoder layer~\citep{vaswani2017attention}: multi-head self-attention over $\zv_a$, followed by multi-head inter-attention over $\xv$, followed by a position-wise feed-forward network.
The input $\zv_a$ is fed into this layer in one pass, without causal masking.

As in \citet{dinh2016density}, the $\mathrm{split}()$ function splits $\zv$ the input tensor into two halves, while the $\mathrm{concat}$ operation performs the corresponding reverse concatenation operation.
In our architecture, three types of split functions are used, based on the split dimension and pattern.
Figure~\ref{fig:flowstep} (b) illustrates the three splitting types.
The first type of split groups $\zv$ along the time dimension on alternate indices.
In this case, FlowSeq mainly models the interactions between time-steps.
The second and third types of splits perform on the feature dimension, with continuous and alternate patterns, respectively.
For each type of split, we alternate $\zv_a$ and $\zv_b$ to increase the flexibility of the split function.
Different types of affine coupling layers alternate in the flow, similar to the linear layers.

\paragraph{Multi-scale Architecture.}
We follow \citet{dinh2016density} in implementing a multi-scale architecture using the squeezing operation on the feature dimension, which has been demonstrated helpful for training deep flows. 
Formally, each scale is a combination of several steps of the flow (see Figure~\ref{fig:flowstep} (a)).
After each scale, the model drops half of the dimensions with the third type of split in Figure~\ref{fig:flowstep} (b) to reduce computational and memory cost, outputting the tensor with shape $[T \times \frac{d}{2}]$.
Then the squeezing operation transforms the $T \times \frac{d}{2}$ tensor into an $\frac{T}{2} \times d$ one as the input of the next scale. 
We pad each sentence with \textsf{EOS} tokens to ensure $T$ is divisible by $2$.
The right component of Figure~\ref{fig:model} illustrates the multi-scale architecture.

\subsection{Predicting Target Sequence Length}
\label{sec:len}
In autoregressive seq2seq models, it is natural to determine the length of the sequence dynamically by simply predicting a special \textsf{EOS} token.
However, for FlowSeq to predict the entire sequence in parallel, it needs to know its length in advance to generate the latent sequence $\zv$. 
Instead of predicting the absolute length of the target sequence, we predict the length difference between source and target sequences using a classifier with a range of $[-20, 20]$.
Numbers in this range are predicted by max-pooling the source encodings into a single vector,%
\footnote{We experimented with other methods such as mean-pooling or taking the last hidden state and found no major difference in our experiments}
running this through a linear layer, and taking a softmax.
This classifier is learned jointly with the rest of the model.

\vspace{-2mm}
\subsection{Decoding Process}
At inference time, the model needs to identify the sequence with the highest conditional probability by marginalizing over all possible latent variables (see Eq.~\eqref{eq:non_auto}), which is intractable in practice.
We propose three approximating decoding algorithms to reduce the search space.
\vspace{-2mm}
\paragraph{Argmax Decoding.} 
Following \citet{gu2018non}, one simple and effective method is to select the best sequence by choosing the highest-probability latent sequence $\zv$:
\begin{displaymath}
\begin{array}{rcl}
\zv^* & = & \argmax\limits_{\zv \in \mathcal{Z}} p_{\theta}(\zv | \xv) \\
\yv^* & = & \argmax\limits_{\yv} P_{\theta} (\yv | \zv^*, \xv)
\end{array}
\vspace{-2mm}
\end{displaymath}
where identifying $\yv^*$ only requires independently maximizing the local probability for each output position (see Eq.~\ref{eq:decoding}).

\paragraph{Noisy Parallel Decoding (NPD).}
A more accurate approximation of decoding, proposed in \citet{gu2018non}, is to draw samples from the latent space and compute the best output for each latent sequence.
Then, a pre-trained autoregressive model is adopted to rank these sequences.
In FlowSeq, different candidates can be generated by sampling different target lengths or different samples from the prior, and both of the strategies can be batched via masks during decoding.
In our experiments, we first select the top $l$ length candidates from the length predictor in \S\ref{sec:len}.
Then, for each length candidate we use $r$ random samples from the prior network to generate output sequences, yielding a total of $l\times r$ candidates.

\paragraph{Importance Weighted Decoding (IWD)}
The third approximating method is based on the \emph{lower bound of importance weighted estimation}~\citep{burda2015importance}.
Similarly to NPD, IWD first draws samples from the latent space and computes the best output for each latent sequence.
Then, IWD ranks these candidate sequences with $K$ importance samples:
\begin{displaymath}
\begin{array}{rcl}
\zv_i & \sim &  p_{\theta}(\zv|\xv), \forall i = 1, \ldots, N \\
\hat{\yv}_i & = &  \argmax\limits_{\yv} P_{\theta} (\yv | \zv_i, \xv) \\
\zv_{i}^{(k)} & \sim & q_{\phi}(\zv | \hat{\yv}_i, \xv), \forall k = 1, \ldots, K \\
P(\hat{\yv}_i | \xv) & \approx & \frac{1}{K} \sum\limits_{k=1}^{K} \frac{P_{\theta}(\hat{\yv}_i |\zv_{i}^{(k)},  \xv) p_{\theta}(\zv_{i}^{(k)}|\xv)}{q_{\phi}(\zv_{i}^{(k)} | \hat{\yv}_i, \xv)}
\end{array}
\end{displaymath}
IWD does not rely on a separate pre-trained model, though it significantly slows down the decoding speed.
The detailed comparison of these three decoding methods is provided in \S\ref{subsec:result}.

\subsection{Discussion}
Different from the architecture proposed in \citet{ziegler2019latent}, the architecture of FlowSeq is not using any autoregressive flow \citep{kingma1606improving,papamakarios2017masked}, yielding a truly non-autoregressive model with both efficient density estimation and generation.
Note that FlowSeq remains non-autoregressive even if we use an RNN in the architecture because RNN is only used to encode a complete sequence of codes and all the input tokens can be fed into the RNN in parallel.
This makes it possible to use highly-optimized implementations of RNNs such as those provided by cuDNN.\footnote{https://devblogs.nvidia.com/optimizing-recurrent-neural-networks-cudnn-5/}
Thus while RNNs do experience some drop in speed, it is less extreme than that experienced when using autoregressive models. 

\section{Experiments}
\subsection{Experimental Setups}
\paragraph{Translation Datasets}
We evaluate FlowSeq on three machine translation benchmark datasets: WMT2014 DE-EN (around 4.5M sentence pairs), WMT2016 RO-EN (around 610K sentence pairs) and a smaller dataset IWSLT2014 DE-EN (around 150K sentence pairs) \cite{cettolo2012wit3}. We use scripts from fairseq \cite{ott2019fairseq} to preprocess WMT2014 and IWSLT2014, where the preprocessing steps follow  \citet{vaswani2017attention} for WMT2014. We use the data provided by \citet{lee2018deterministic} for WMT2016. For both WMT datasets, the source and target languages share the same set of subword embeddings while for IWSLT2014 we use separate embeddings.
During training, we filter out sentences longer than $80$ for WMT dataset and $60$ for IWSLT, respectively.

\begin{table}[t]
\centering
\resizebox{1.0\columnwidth}{!}
{%
\begin{tabular}{lccccc}
\toprule
 & \multicolumn{2}{c}{\textbf{WMT2014}} & \multicolumn{2}{c}{\textbf{WMT2016}} & \textbf{IWSLT2014}\\
\textbf{ Models} & \textbf{EN-DE} & \textbf{DE-EN} & \textbf{EN-RO} & \textbf{RO-EN} & \textbf{DE-EN} \\
\midrule\midrule
\multicolumn{6}{c}{Raw Data} \\
\midrule
CMLM-base & 10.88 & -- & 20.24 & -- & -- \\
LV NAR & 11.80 & -- & -- & -- & -- \\
\midrule
FlowSeq-base & 18.55 & 23.36 & 29.26 & 30.16 & \textbf{24.75} \\
FlowSeq-large & \textbf{20.85} & \textbf{25.40} & \textbf{29.86} & \textbf{30.69} & -- \\
\midrule\midrule
\multicolumn{6}{c}{Knowledge Distillation} \\ 
\midrule
NAT-IR & 13.91 & 16.77 & 24.45 & 25.73 & 21.86 \\
CTC Loss & 17.68 & 19.80 & 19.93 & 24.71 & -- \\
NAT w/ FT & 17.69 & 21.47 & 27.29 & 29.06 & 20.32 \\
NAT-REG & 20.65 & 24.77 & -- & -- & 23.89 \\
CMLM-small & 15.06 & 19.26 & 20.12 & 20.36 & -- \\
CMLM-base & 18.12 & 22.26 & 23.65 & 22.78 & -- \\
\midrule
FlowSeq-base & 21.45 & 26.16 & 29.34 & 30.44 & \textbf{27.55} \\
FlowSeq-large & \textbf{23.72} & \textbf{28.39} & \textbf{29.73} & \textbf{30.72} & -- \\
\bottomrule
\end{tabular}
}
\caption{BLEU scores on three MT benchmark datasets for FlowSeq with argmax decoding and baselines with purely non-autoregressive decoding methods.  
The first and second block are results of models trained w/w.o. knowledge distillation, respectively.}
\label{tab:pure}
\vspace{-4mm}
\end{table}

\begin{table}[t]
\centering
\resizebox{1.0\columnwidth}{!}
{%
\begin{tabular}{lcccc}
\toprule
 & \multicolumn{2}{c}{\textbf{WMT2014}} & \multicolumn{2}{c}{\textbf{WMT2016}} \\
\textbf{ Models} & \textbf{EN-DE} & \textbf{DE-EN} & \textbf{EN-RO} & \textbf{RO-EN} \\
\midrule\midrule
\multicolumn{5}{c}{Autoregressive Methods} \\
\midrule
Transformer-base & 27.30 &  -- &  -- & -- \\
Our Implementation & 27.16 & 31.44 & 32.92 & 33.09 \\
\midrule\midrule
\multicolumn{5}{c}{Raw Data} \\
\midrule
CMLM-base (refinement 4) & 22.06 & -- & 30.89 & -- \\
CMLM-base (refinement 10) & \textbf{24.65} & -- & \textbf{32.53} & -- \\
\midrule
FlowSeq-base (IWD $n=15$) & 20.20 & 24.63 & 30.61 & 31.50 \\
FlowSeq-base (NPD $n=15$) & 20.81 & 25.76 & 31.38 & 32.01 \\
FlowSeq-base (NPD $n=30$) & 21.15 & 26.04 & 31.74 & 32.45 \\
\hdashline
FlowSeq-large (IWD $n=15$) & 22.94 & 27.16 & 31.08 & 32.03 \\
FlowSeq-large (NPD $n=15$) & 23.14 & 27.71 & 31.97 & 32.46 \\
FlowSeq-large (NPD $n=30$) & 23.64 & \textbf{28.29} & 32.35 & \textbf{32.91} \\
\midrule\midrule
\multicolumn{5}{c}{Knowledge Distillation} \\ 
\midrule
NAT-IR (refinement 10) & 21.61 &  25.48 & 29.32 & 30.19 \\
NAT w/ FT (NPD $n=10$) & 18.66 & 22.42 & 29.02 & 31.44 \\
NAT-REG (NPD $n=9$) & 24.61 & 28.90 & -- & -- \\
LV NAR (refinement 4) & 24.20 & -- & -- & -- \\
CMLM-small (refinement 10) & 25.51 & 29.47 & 31.65 & 32.27 \\
CMLM-base (refinement 10) & \textbf{26.92} & \textbf{30.86} & \textbf{32.42} & \textbf{33.06} \\
\midrule
FlowSeq-base (IWD $n=15$) & 22.49 & 27.40 & 30.59 & 31.58 \\
FlowSeq-base (NPD $n=15$) & 23.08 & 28.07 & 31.35 & 32.11 \\
FlowSeq-base (NPD $n=30$) & 23.48 & 28.40 & 31.75 & 32.49 \\
\hdashline
FlowSeq-large (IWD $n=15$) & 24.70  & 29.44 & 31.02 & 31.97 \\
FlowSeq-large (NPD $n=15$) & 25.03 & 30.48 & 31.89 & 32.43 \\
FlowSeq-large (NPD $n=30$) & 25.31 & 30.68 & 32.20 & 32.84 \\
\bottomrule
\end{tabular}
}
\caption{BLEU scores on two WMT datasets of models using advanced decoding methods. 
The first block are autoregressive Transformer-base~\citep{vaswani2017attention}.
The second and third blocks are results of models trained w/w.o. knowledge distillation, respectively.
$n=l\times r$ is the total number of rescoring candidates.}
\label{tab:refine}
\vspace{-4mm}
\end{table}

\paragraph{Modules and Hyperparameters}
We implement the encoder, decoder and posterior networks with standard (unmasked) Transformer layers~\citep{vaswani2017attention}.
For WMT datasets, we use 8 attention heads, the encoder consists of 6 layers, and the decoder and posterior are composed of 4 layers.
For IWSLT, we use 4 attention heads, the encoder has 5 layers, and decoder and posterior have 3 layers.
The prior flow consists of 3 scales with the number of steps $[48, 48, 16]$ from bottom to top.
To dissect the impact of model dimension on translation quality and speed, we perform experiments on two versions of FlowSeq with $d_{model}/d_{hidden} = 256/512$ (base) and $d_{model}/d_{hidden} = 512/1024$ (large).
More model details are provided in Appendix~\ref{appendix:model}.

\paragraph{Optimization}
Parameter optimization is performed with the Adam optimizer~\citep{kingma2014adam} with $\beta=(0.9, 0.999)$,  $\epsilon=1e^{-8}$ and AMSGrad~\citep{iclr2018amsgrad}. 
Each mini-batch consist of $2048$ sentences.
The learning rate is initialized to $5e-4$, and exponentially decays with rate $0.999995$.
The gradient clipping cutoff is $1.0$.
For all the FlowSeq models, we apply $0.1$ label smoothing~\citep{vaswani2017attention} and averaged the 5 best checkpoints to create the final model.

At the beginning of training, the posterior network is randomly initialized, producing noisy supervision to the prior.
To mitigate this issue, we first set the weight of the $\mathrm{KL}$ term in the ELBO to zero for 30,000 updates to train the encoder, decoder and posterior networks.
Then the $\mathrm{KL}$ weight linearly increases to one for another 10,000 updates, which we found essential to accelerate training and achieve stable performance.

\paragraph{Knowledge Distillation}
Previous work on non-autoregressive generation \citep{gu2018non,constant2019} has used translations produced by a pre-trained autoregressive NMT model as the training data, noting that this can significantly improve the performance.
We analyze the impact of distillation in \S~\ref{subsec:result}.

\begin{figure*}[t]
  \centering
  \begin{minipage}[t]{1.0\textwidth}
  \subfloat[batch size]{
  \includegraphics[width=0.49\textwidth]{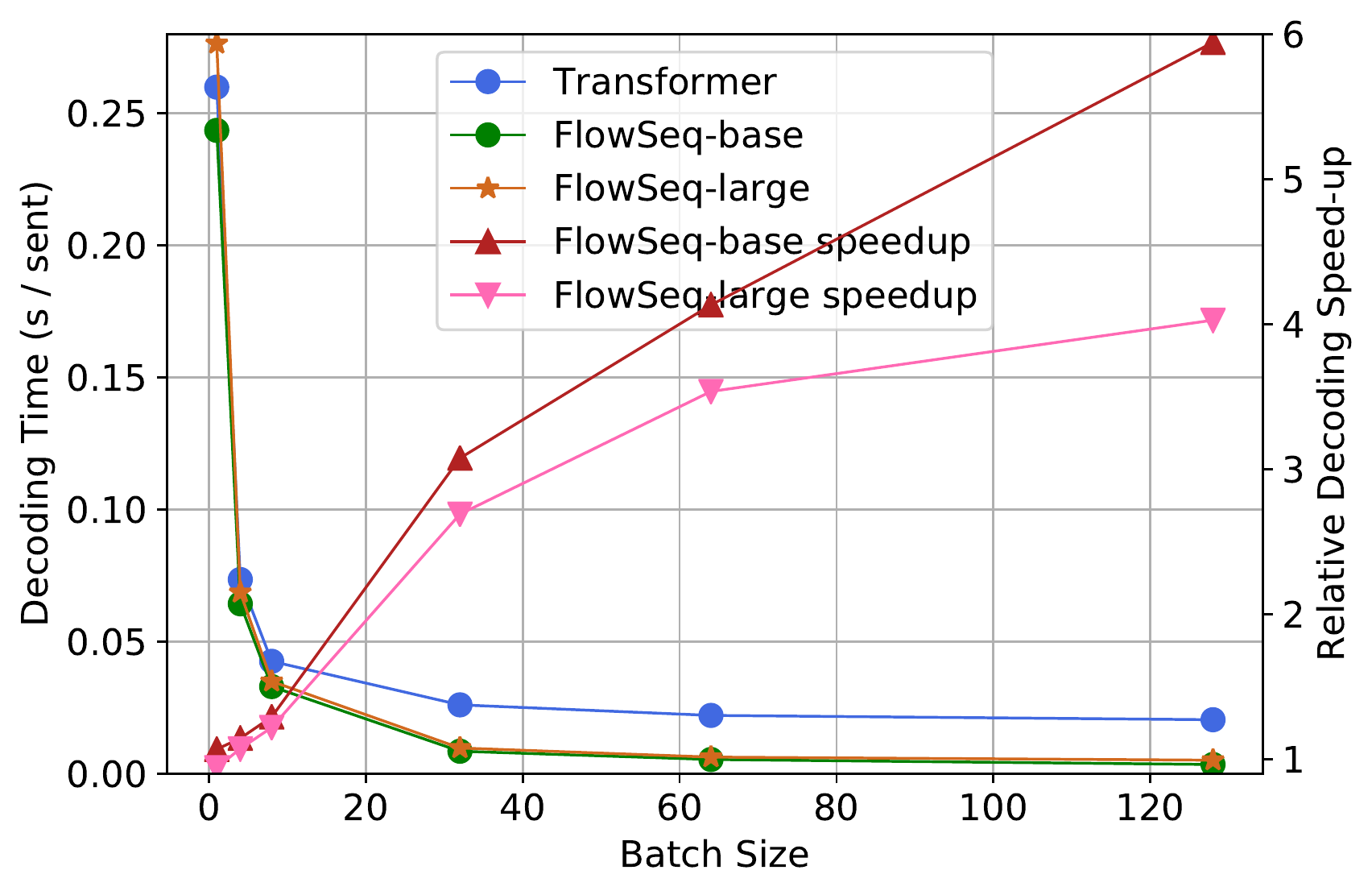}
  \label{fig:batch}
  }
  \subfloat[target length]{
  \includegraphics[width=0.49\textwidth]{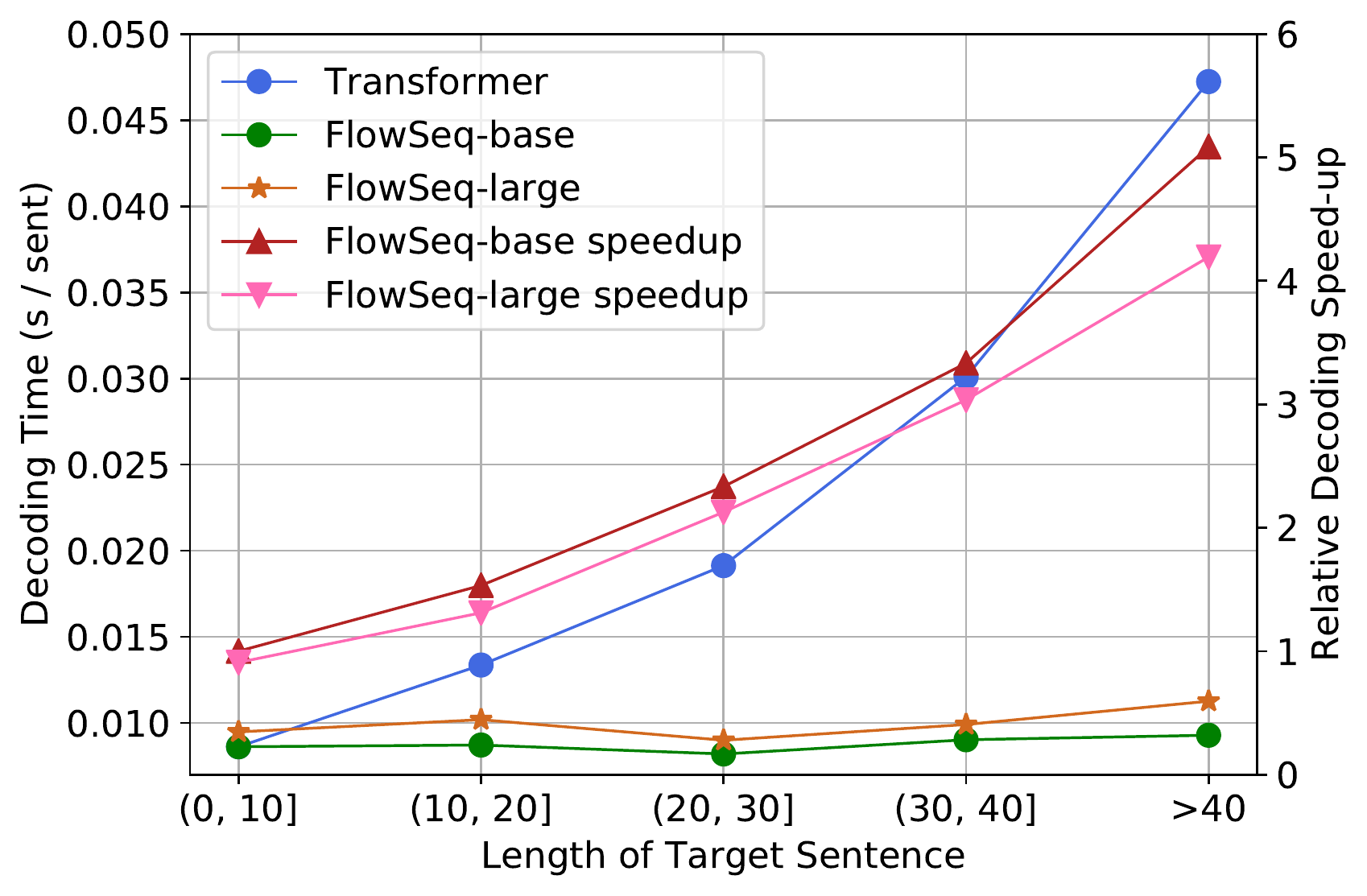}
  \label{fig:length}
  }
  \end{minipage}
  \caption{The decoding speed of the Transformer (batched, beam size 5) and FlowSeq on WMT14 EN-DE test set (a) w.r.t.~different batch sizes (b) bucketed by different target sentence lengths (batch size 32).}
  \label{fig:speed}
  \vspace{-4mm}
\end{figure*}

\subsection{Main Results}\label{subsec:result}
We first conduct experiments to compare the performance of FlowSeq with strong baseline models, including NAT w/ Fertility~\cite{gu2018non}, NAT-IR~\cite{lee2018deterministic}, NAT-REG~\cite{wang2019non}, LV NAR~\cite{shu2019latent}, CTC Loss~\cite{libovicky2018end}, and CMLM~\cite{constant2019}.

Table~\ref{tab:pure} provides the BLEU scores of FlowSeq with argmax decoding, together with baselines with purely non-autoregressive decoding methods that generate output sequence in one parallel pass.
The first block lists results of models trained on raw data, while the second block shows results using knowledge distillation.
Without using knowledge distillation, the FlowSeq base model achieves significant improvements (more than $9$ BLEU points) over the baselines.
This demonstrates the effectiveness of FlowSeq in modeling complex interdependences in the target languages.

Regarding the effect of knowledge distillation, we can mainly obtain two observations: i)
Similar to the findings in previous work, knowledge distillation still benefits the translation quality of FlowSeq. 
ii) Compared to previous models, the benefit of knowledge distillation for FlowSeq is less significant, yielding less than $3$ BLEU improvement on WMT2014 DE-EN corpus, and even no improvement on WMT2016 RO-EN corpus.
We hypothesize that the reason for this is that FlowSeq's stronger model is more robust against multi-modality, making it less necessary to rely on knowledge distillation.

Table~\ref{tab:refine} illustrates the BLEU scores of FlowSeq and baselines with advanced decoding methods such as iterative refinement, IWD and NPD rescoring.
The first block in Table~\ref{tab:refine} includes the baseline results from autoregressive Transformer.
For the sampling procedure in IWD and NPD, we sampled from a reduced-temperature model~\citep{kingma2018glow} to obtain high-quality samples.
We vary the temperature within $\{0.1, 0.2, 0.3, 0.4, 0.5, 1.0\}$ and select the best temperature based on the performance on development sets.
The analysis of the impact of sampling temperature and other hyper-parameters on samples is shown in \S~\ref{subsec:rescore}.
For FlowSeq, NPD obtains better results than IWD, showing that FlowSeq still falls behind the autoregressive Transformer on modeling the distributions of target languages.
Compared with CMLM \citep{constant2019} with $10$ iterations of refinement, which is a contemporaneous work that achieves state-of-the-art translation performance, FlowSeq obtains competitive performance on both WMT2014 and WMT2016 corpora, with only slight degradation in translation quality.
Notably we did not attempt to perform iterative refinement, but there is nothing that makes FlowSeq inherently incompatible with refinement -- we leave connecting the two techniques to future work.

\subsection{Analysis on Decoding Speed}
In this section, we compare the decoding speed (measured in average time in seconds required to decode one sentence) of FlowSeq at test time with that of the autoregressive Transformer model. 
We use the test set of WMT14 EN-DE for evaluation and all experiments are conducted on a single NVIDIA TITAN X GPU.
\vspace{-1mm}
\paragraph{How does batch size affect the decoding speed?}
First, we investigate how different decoding batch size can affect the decoding speed. We vary the decoding batch size within $\{1, 4, 8, 32, 64, 128\}$. Figure.~\ref{fig:batch} shows that for both FlowSeq and the autoregressive Transformer decoding is faster when using a larger batch size. However, FlowSeq has much larger gains in the decoding speed w.r.t.~the increase in batch size, gaining a speed up of 594\% of the base model and 403\% of the large model when using a batch size of 128. We hypothesize that this is because the operations in FlowSeq are more friendly to batching while the incremental nature of left-to-right search in the autoregressive model is less efficient in benefiting from batching.

\paragraph{How does sentence length affect the decoding speed?}
Next, we examine if sentence length is a major factor affecting the decoding speed. We bucket the test data by the target sentence length. From Fig.~\ref{fig:length}, we can see that as the sentence length increases, FlowSeq achieves almost a constant decoding time while the autoregressive Transformer has a linearly increasing decoding time. The relative decoding speed of FlowSeq versus the Transformer linearly increases as the sequence length increases. The potential of decoding long sequences with constant time is an attractive property of FlowSeq.

\begin{figure}[tb]
  \centering
  \includegraphics[width=0.99\columnwidth]{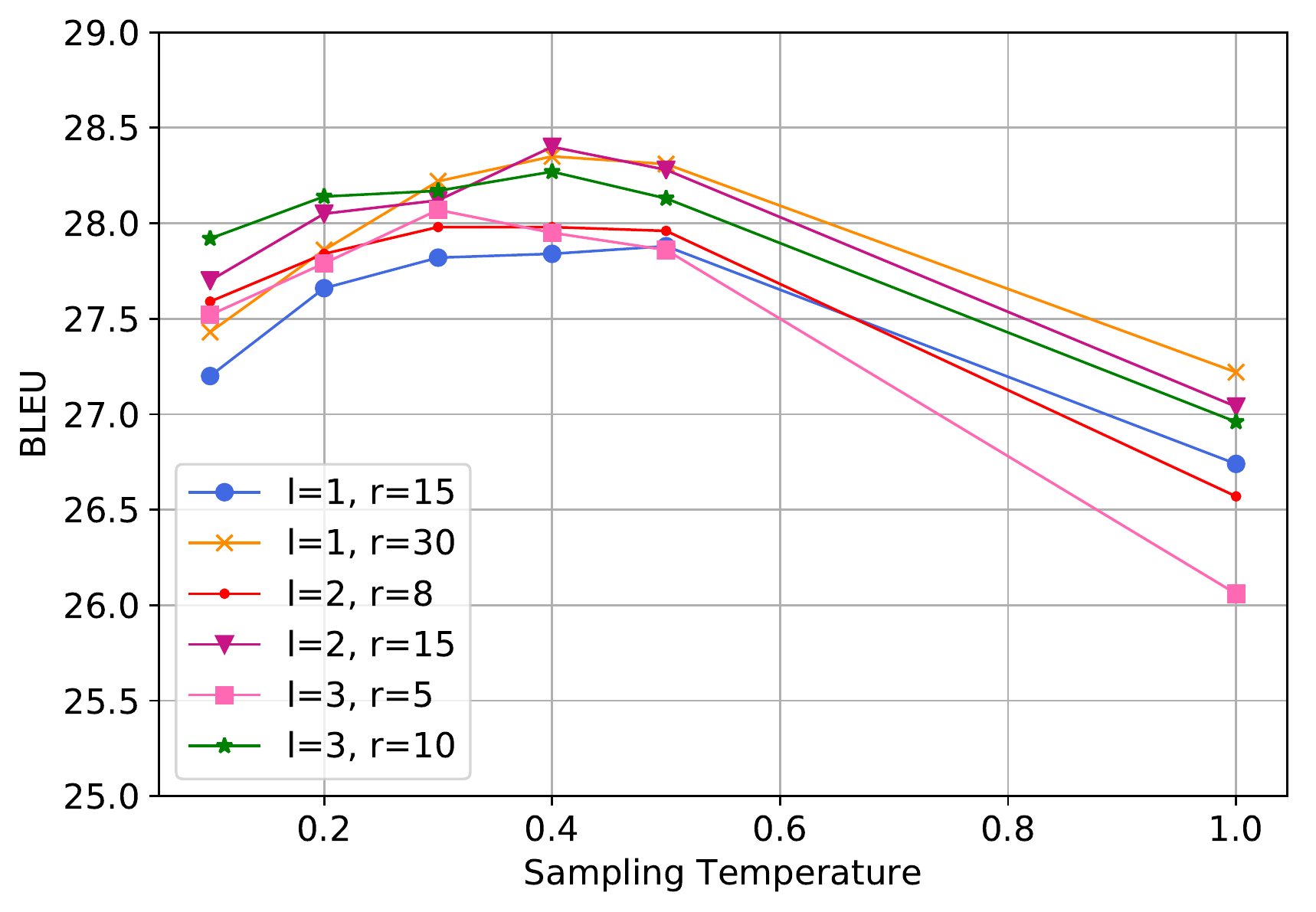}
  \caption{Impact of sampling hyperparameters on the rescoring BLEU on the dev set of WMT14 DE-EN. Experiments are performed with FlowSeq-base trained with distillation data. $l$ is the number of length candidates. $r$ is the number of samples for each length.} \label{fig:rerank}
  \vspace{-4mm}
\end{figure}
\subsection{Analysis of Rescoring Candidates}\label{subsec:rescore}
In Fig.~\ref{fig:rerank}, we analyze how different sampling hyperparameters affect the performance of rescoring. First, we observe that the number of samples $r$ for each length is the most important factor. The performance is always improved with a larger sample size. Second, a larger number of length candidates does not necessarily increase the rescoring performance. Third, we find that a larger sampling temperature (0.3 - 0.5) can increase the diversity of translations and leads to better rescoring BLEU. However, the latent samples become noisy when a large temperature (1.0) is used.

\subsection{Analysis of Translation Diversity}
\begin{figure}[tb]
  \centering
  \includegraphics[width=0.97\columnwidth]{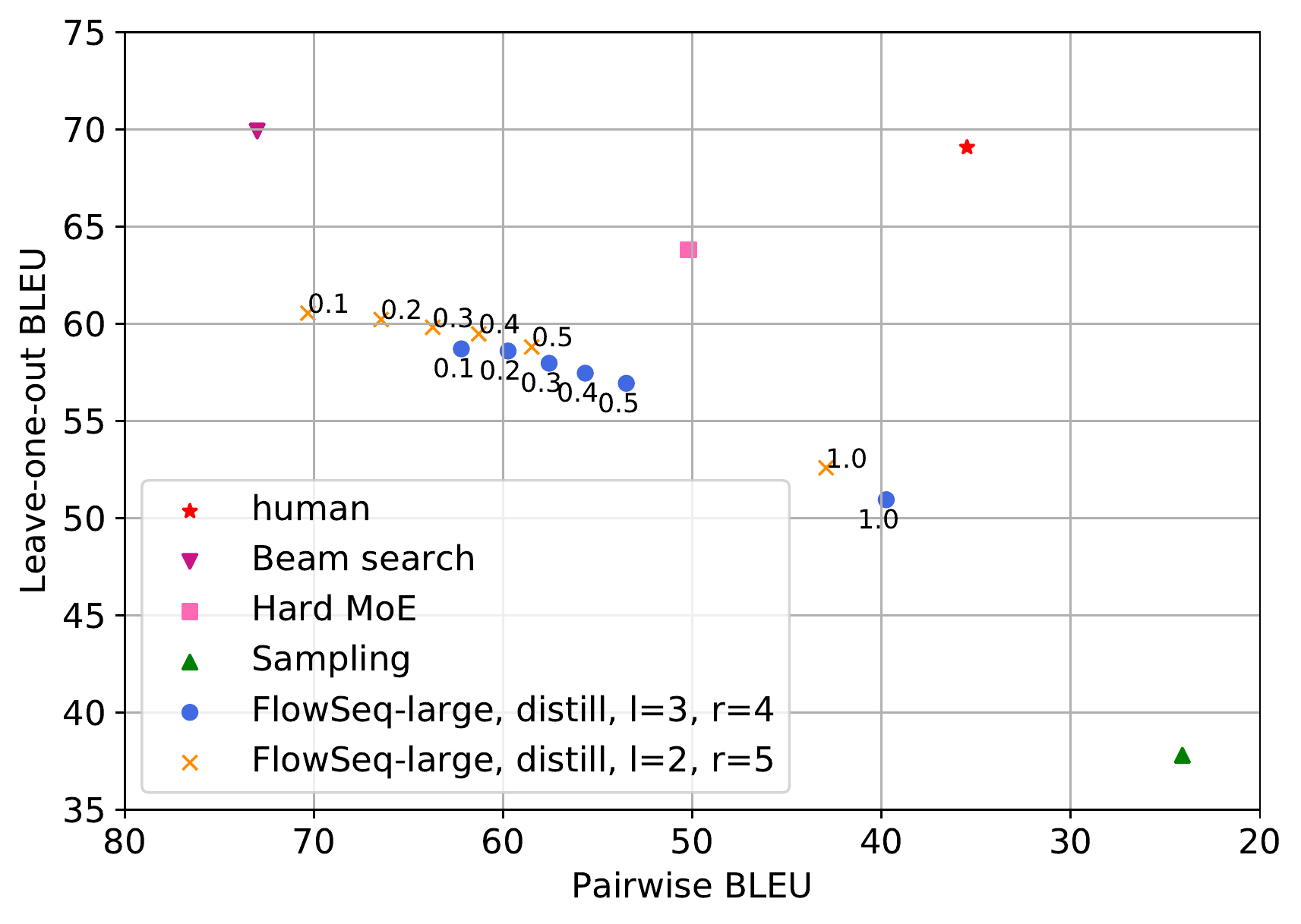}
  \caption{Comparisons of FlowSeq with human translations, beam search and sampling results of Transformer-base, and mixture-of-experts model (Hard MoE \cite{shen2019mixture}) on the averaged leave-one-out BLEU score v.s pairwise-BLEU in descending order.} \label{fig:diverse}
  \vspace{-4mm}
\end{figure}
Following \citet{he2018sequence} and \citet{shen2019mixture}, we analyze the output diversity of FlowSeq. They proposed pairwise-BLEU and BLEU computed in a leave-one-out manner to calibrate the diversity and quality of translation hypotheses. A lower pairwise-BLEU score implies a more diverse hypothesis set. And a higher BLEU score implies a better translation quality. We experiment on a subset of the test set of WMT14-ENDE with ten references for each sentence \cite{ott2018analyzing}.
In Fig.~\ref{fig:diverse}, we compare FlowSeq with other multi-hypothesis generation methods (ten hypotheses each sentence) to analyze how well the generation outputs of FlowSeq are in terms of diversity and quality. The right corner area of the figure indicates the ideal generations: high diversity and high quality. While FlowSeq still lags behind the autoregressive generation, by increasing the sampling temperature it provides a way of generating more diverse outputs while keeping the translation quality almost unchanged. More analysis of translation outputs and detailed results are provided in the Appendix~\ref{appendix:sample} and \ref{appendix:deversity}.

\vspace{-1mm}
\section{Conclusion}
We propose FlowSeq, an efficient and effective model for non-autoregressive sequence generation by using generative flows.
One potential direction for future work is to leverage iterative refinement techniques such as masked language models to further improve translation quality. 
Another exciting direction is to, theoretically and empirically, investigate the latent space in FlowSeq, hence providing deeper insights into the model, and allowing for additional applications such as controllable text generation.

\vspace{-1mm}
\section*{Acknowledgments}
Xuezhe MA was supported in part by DARPA grant FA8750-18-2-0018 funded under the AIDA
program and Chunting Zhou was supported by DARPA grant HR0011-15-C-0114 funded under the LORELEI program.
Any opinions, findings, and conclusions expressed in this material are those of the authors and do not necessarily reflect the views of DARPA.
The authors thank Amazon for their gift of AWS cloud credits and anonymous reviewers for their helpful suggestions.

\bibliography{emnlp-ijcnlp-2019}
\bibliographystyle{acl_natbib}

\newpage
\appendix
\onecolumn
\begin{center}
  {\huge {\bf Appendix: FlowSeq}}  
\end{center}

\section{Flow Layers}\label{appendix:flow}
\paragraph{ActNorm}
\begin{displaymath}
\zv_{t}' = \mathbf{s} \odot \zv_{t} + \mathbf{b}.
\end{displaymath}
Log-determinant:
\begin{displaymath}
T \cdot \mathrm{sum}(\log |\mathbf{s}|)
\end{displaymath}

\paragraph{Invertible Linear}
\begin{displaymath}
\zv_{t}' = \zv_{t} \mathbf{W},
\end{displaymath}
Log-determinant:
\begin{displaymath}
T \cdot h \cdot \log |\mathrm{det} (\mathbf{W})|
\end{displaymath}
where $h$ is the number of heads.

\paragraph{Affine Coupling}
\begin{displaymath}
\begin{array}{rcl}
\zv_a, \zv_b & = & \mathrm{split}(\zv) \\
\zv_a' & = & \zv_a \\
\zv_b' & = & \mathrm{s}(\zv_a, \xv) \odot \zv_b + \mathrm{b}(\zv_a, \xv) \\
\zv' & = & \mathrm{concat}(\zv_a', \zv_b'),
\end{array}
\end{displaymath}
Log-determinant:
\begin{displaymath}
\mathrm{sum}(\log |\mathbf{s}|)
\end{displaymath}

\section{Model Details}
\label{appendix:model}
\begin{table}[!ht]
\centering
\begin{tabular}[t]{ccc}
\toprule
\textbf{Model} & \textbf{Dimensions (Model/Hidden)} & \textbf{\#Params} \\
Transformer-base & 512/2048 & 65M \\
Transformer-large & 2014/4096 & 218M \\
\midrule
FlowSeq-base & 256/512 & 73M \\
FlowSeq=large & 512/2014 & 258M \\
\bottomrule
\end{tabular}
\caption{Comparison of model size in our experiments.}
\label{tab:model_size}
\end{table}

\newpage
\section{Analysis of training dynamics}
\begin{figure}[h]
  \begin{minipage}[t]{1.0\columnwidth}
  \centering
  \subfloat[training loss]{
  \includegraphics[width=0.49\textwidth]{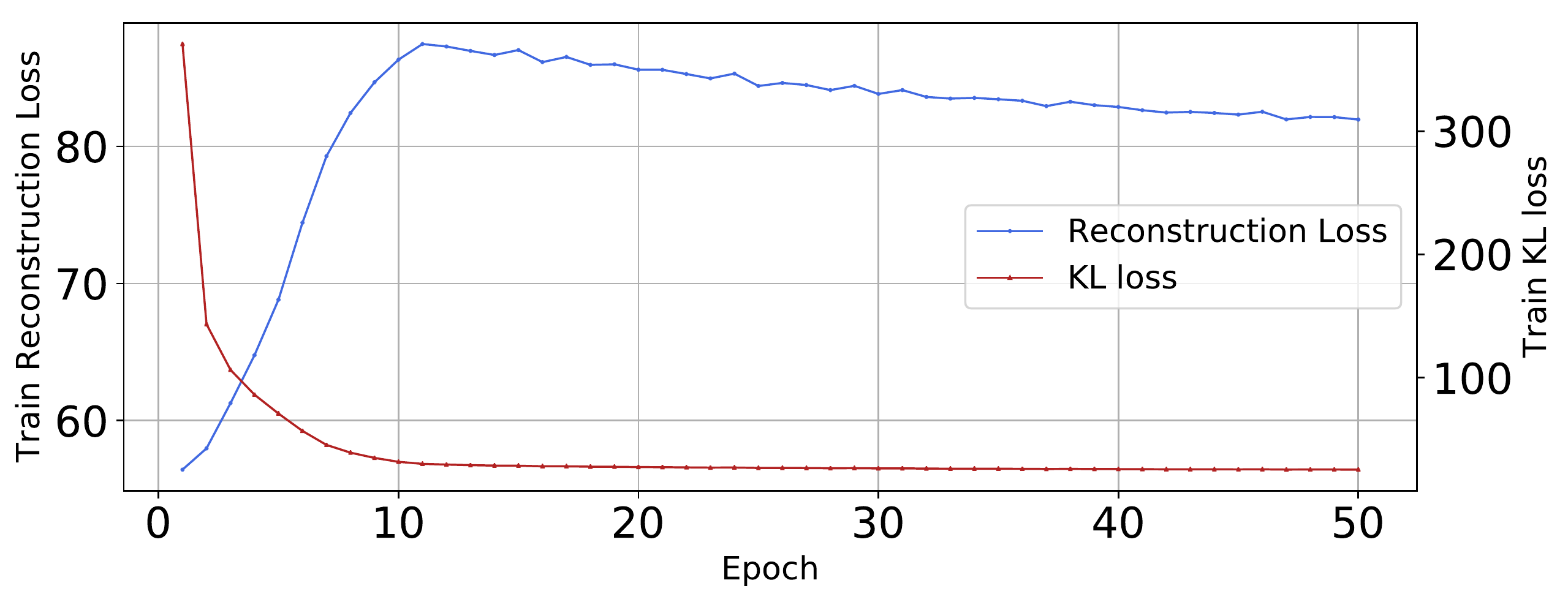}
  \label{fig:train:loss}
  }
  \\
  \subfloat[dev loss]{
  \includegraphics[width=0.49\textwidth]{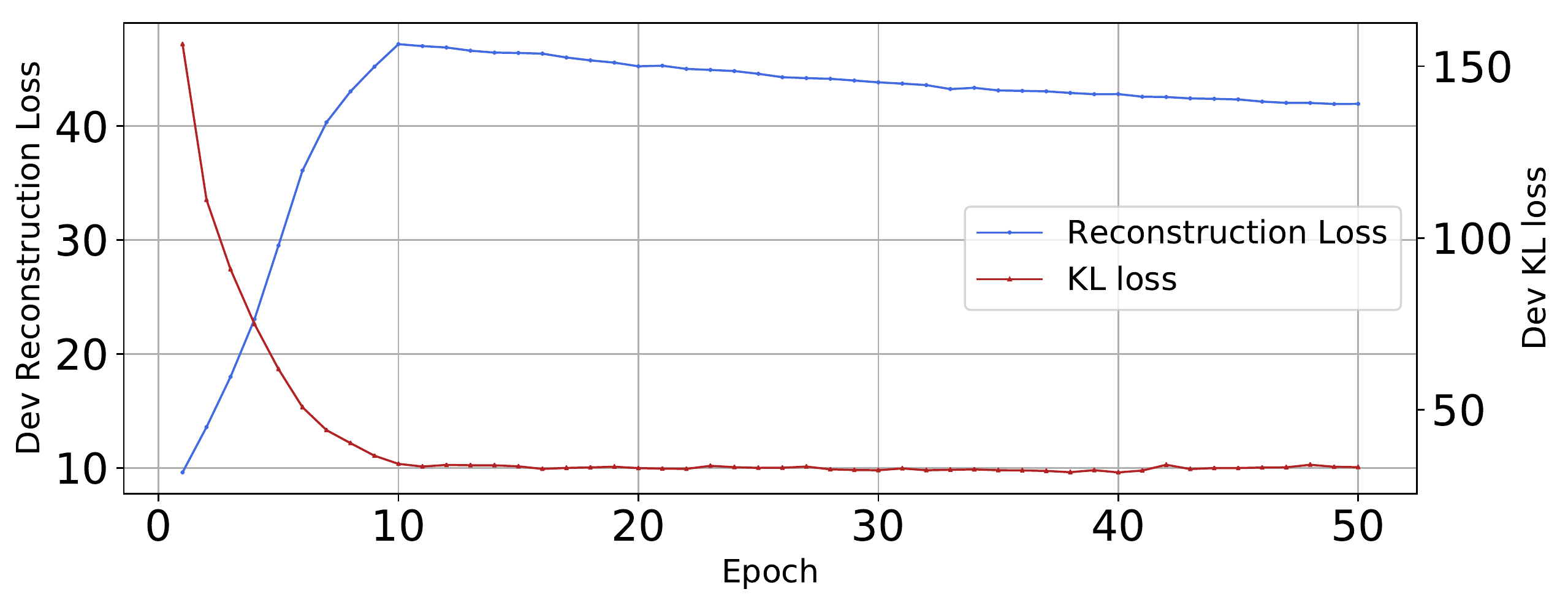}
  \label{fig:dev:loss}
  }
  \subfloat[dev BLEU]{
  \includegraphics[width=0.49\textwidth]{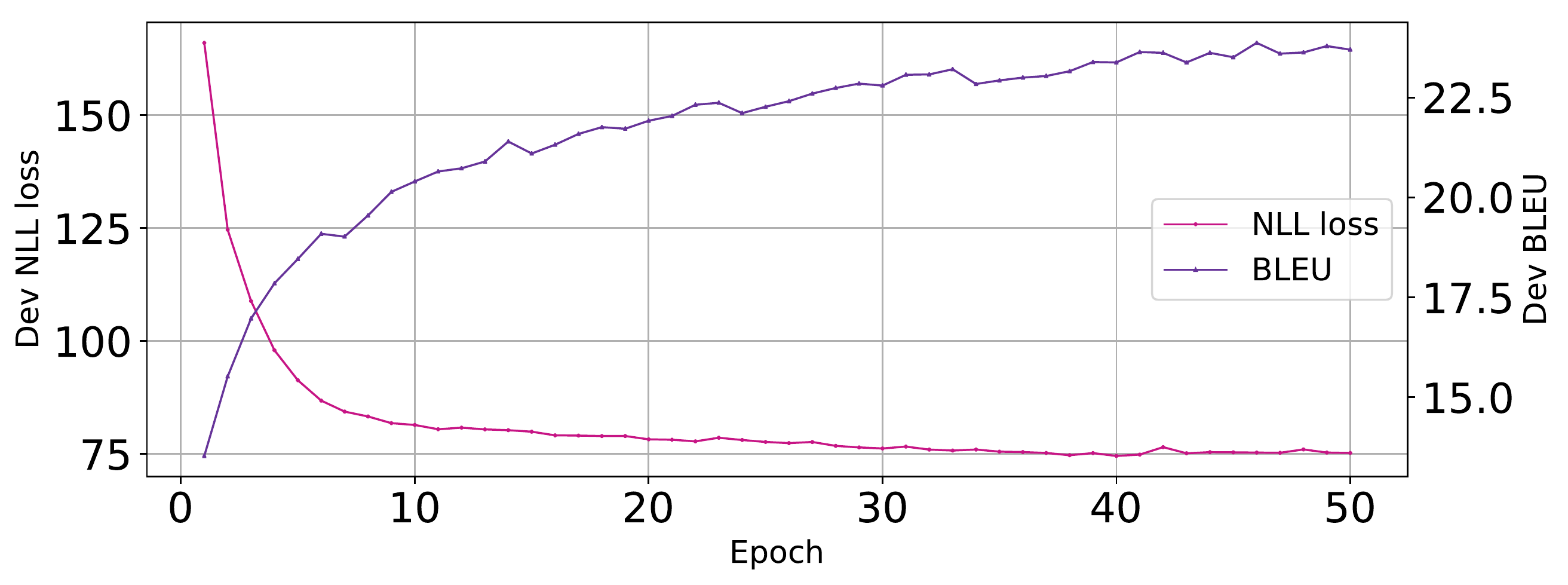}
  \label{fig:dev:bleu}
  }
  \end{minipage}
  \caption{Training dynamics.}
  \label{fig:dynamic}
\end{figure}
In Fig. \ref{fig:dynamic}, we plot the train and dev loss together with dev BLEU scores for the first 50 epochs. We can see that the reconstruction loss is increasing at the initial stage of training, then starts to decrease when training with full KL loss.
In addition, we observed that FlowSeq does not suffer the KL collapse problem~\citep{bowman2015generating,iclr2019:ma:mae}.
This is because the decoder of FlowSeq is non-autogressive, with latent variable $\zv$ as the only input.

\newpage
\section{Analysis of Translation Results}\label{appendix:sample}
\begin{table*}[ht!]
\centering
\begin{tabular}{p{2.5cm}p{12cm}}
\toprule

\textcolor{myblue}{Source} & \textcolor{myblue}{Grundnahrungsmittel gibt es schließlich überall und jeder Supermarkt hat mittlerweile Sojamilch und andere Produkte.} \\
\textcolor{mypink}{Ground Truth} & 
\textcolor{mypink}{There are basic foodstuffs available everywhere , and every supermarket now has soya milk and other products.} \\
Sample 1 &  
After all, there are basic foods everywhere and every supermarket now has soya amch and other products. \\
Sample 2 &  After all, the food are available everywhere everywhere and every supermarket has soya milk and other products. \\
Sample 3 & After all, basic foods exist everywhere and every supermarket has now had soy milk and other products. \\
\midrule
\textcolor{myblue}{Source} & \textcolor{myblue}{Es kann nicht erkl\"{a}ren, weshalb die National Security Agency Daten über das Privatleben von Amerikanern sammelt und warum Whistleblower bestraft werden, die staatliches Fehlverhalten offenlegen.}\\
\textcolor{mypink}{Ground Truth} & \textcolor{mypink}{And, most recently, it cannot excuse the failure to design a simple website more than three years since the Affordable Care Act was signed into law.}\\
Sample 1 &  And recently, it cannot apologise for the inability to design a simple website in the more than three years since the adoption of Affordable Care Act.\\
Sample 2 &  And recently, it cannot excuse the inability to design a simple website in more than three years since the adoption of Affordable Care Act.\\
Sample 3 & Recently, it cannot excuse the inability to design a simple website in more than three years since the Affordable Care Act has passed. \\
\midrule
\textcolor{myblue}{Source} & \textcolor{myblue}{Doch wenn ich mir die oben genannten Beispiele ansehe, dann scheinen sie weitgehend von der Regierung selbst gew\"{a}hlt zu sein.}\\
\textcolor{mypink}{Ground Truth} & \textcolor{mypink}{Yet, of all of the examples that I have listed above, they largely seem to be of the administration's own choosing.}
\\
Sample 1 & However, when I look at the above mentioned examples, they seem to be largely elected by the government itself. \\
Sample 2 &  But if I look at the above mentioned examples, they seem to have been largely elected by the government itself.\\
Sample 3 &  But when I look at the above examples, they seem to be largely chosen by the government itself. \\
\midrule
\textcolor{myblue}{Source} & \textcolor{myblue}{Damit wollte sie auf die Gefahr von noch gr\"{o}ßeren Ruinen auf der Schweizer Wiese hinweisen - sollte das Riesenprojekt eines Tages scheitern.} \\
\textcolor{mypink}{Ground Truth} & \textcolor{mypink}{In so doing they wanted to point out the danger of even bigger ruins on the Schweizer Wiese - should the huge project one day fail.} \\
Sample 1 &  In so doing, it wanted to highlight the risk of even greater ruins on the Swiss meadow - the giant project should fail one day. \\
Sample 2 & In so doing, it wanted to highlight the risk of even greater ruins on the Swiss meadow - if the giant project fail one day. \\
Sample 3 &  In doing so, it wanted point out the risk of even greater ruins on the Swiss meadow - the giant project would fail one day. \\
\bottomrule
\end{tabular}
\caption{Examples of translation outputs from FlowSeq-base with sampling hyperparameters $l=3, r=10, \tau=0.4$ on WMT14-DEEN.}
\label{tab:example}
    \vspace{-2mm}
\end{table*}
In Tab. \ref{tab:example}, we present randomly picked translation outputs from the test set of WMT14-DEEN. For each German input sentence, we pick three hypotheses from 30 samples. We have the following observations: First, in most cases, it can accurately express the meaning of the source sentence, sometimes in a different way from the reference sentence, which cannot be precisely reflected by the BLEU score.
Second, by controlling the sampling hyper-parameters such as the length candidates $l$, the sampling temperature $\tau$ and the number of samples $r$ under each length, FlowSeq is able to generate diverse translations expressing the same meaning.
Third, repetition and broken translations also exist in some cases due to the lack of direct modeling of dependencies between target words.

\section{Results of Translation Diversity}\label{appendix:deversity}
Table~\ref{tab:diverse} shows the detailed results of translation diversity.

\begin{table}[h]
\centering
\resizebox{0.5\columnwidth}{!}
{
\begin{tabular}{llcc}
\toprule
\textbf{Models} & $\tau$ & \textbf{Pairwise BLEU} & \textbf{LOO BLEU} \\
\midrule
Human & -- & 35.48 & 69.07 \\
Sampling & -- & 24.10 & 37.80 \\
Beam Search & -- & 73.00 & 69.90\\
Hard-MoE & -- & 50.02 & 63.80 \\
\midrule
\multirowcell{6}{FlowSeq \\ l=1, r=10} 
& 0.1 & 79.39 & 61.61 \\
& 0.2 & 72.12 & 61.05 \\
& 0.3 & 67.85 & 60.79 \\
& 0.4 & 64.75 & 60.07 \\
& 0.5 & 61.12 & 59.54  \\
& 1.0 & 43.53 & 52.86 \\
\midrule
\multirowcell{6}{FlowSeq \\ l=2, r=5}
& 0.1 & 70.32 & 60.54 \\
& 0.2 & 66.45 & 60.21 \\
& 0.3 & 63.72 & 59.81 \\
& 0.4 & 61.29 & 59.47 \\
& 0.5 & 58.49 & 58.80 \\
& 1.0 & 42.93 & 52.58 \\
\midrule
\multirowcell{6}{FlowSeq \\ l=3, r=4} 
& 0.1 & 62.21 & 58.70 \\
& 0.2 & 59.74 & 58.59\\
& 0.3 & 57.57 & 57.96 \\
& 0.4 & 55.66 & 57.45 \\
& 0.5 & 53.49 & 56.93 \\
& 1.0 & 39.75 & 50.94 \\
\bottomrule
\end{tabular}
}
\caption{Translation diversity results of FlowSeq-large model on WMT14 EN-DE with knowledge distillation.}
\label{tab:diverse}
\vspace{-4mm}
\end{table}

\end{document}